\documentclass[runningheads]{llncs}
\usepackage[T1]{fontenc}
\usepackage{graphicx}
\usepackage{booktabs}
\usepackage{multirow}
\usepackage{amsmath}

\usepackage{xcolor}
\begin{document}
\title{StepChain GraphRAG: Reasoning Over Knowledge Graphs for Multi-Hop Question Answering}
\titlerunning{StepChain GraphRAG}
%

\author{
Tengjun Ni\inst{1} \and
Xin Yuan\inst{2,4} \and
Shenghong Li\inst{2} \and
Kai Wu\inst{1} \and
Ren Ping Liu\inst{1}  \and
Wei Ni\inst{3,4}\thanks{Corresponding author.} \and
Wenjie Zhang\inst{4}
}

\authorrunning{T. Ni et al.}

%

\institute{
University of Technology Sydney, Australia\\
\email{nitengjun2002@gmail.com}\\
\email{Kai.Wu@uts.edu.au}\\
\email{RenPing.Liu@uts.edu.au}
\and
Data61, CSIRO, Australia\\
\email{xin.yuan@ieee.org}\\
\email{shenghong.li@csiro.au}
\and
School of Engineering, Edith Cowan University, Australia\\
\email{Wei.ni@ieee.org}
\and
University of New South Wales, Australia\\
\email{wenjie.zhang@unsw.edu.au}
}

\maketitle              
%


\begin{abstract}
Recent progress in retrieval-augmented generation (RAG) has led to more accurate and interpretable multi-hop question answering (QA). Yet, challenges persist in integrating iterative reasoning steps with external knowledge retrieval. To address this, we introduce StepChain GraphRAG, a framework that unites question decomposition with a Breadth-First Search (BFS) Reasoning Flow for enhanced multi-hop QA. Our approach first builds a global index over the corpus; at inference time, only retrieved passages are parsed on‑the‑fly into a knowledge graph, and the complex query is split into sub‑questions. For each sub-question, a BFS-based traversal dynamically expands along relevant edges, assembling explicit evidence chains without overwhelming the language model with superfluous context. Experiments on MuSiQue, 2WikiMultiHopQA, and HotpotQA show that StepChain GraphRAG achieves state-of-the-art Exact Match and F1 scores. StepChain GraphRAG lifts average EM by 2.57\% and F1 by 2.13\% over the SOTA method, achieving the largest gain on HotpotQA (+4.70\% EM, +3.44\% F1). StepChain GraphRAG also fosters enhanced explainability by preserving the chain-of-thought across intermediate retrieval steps. We conclude by discussing how future work can mitigate the computational overhead and address potential hallucinations from large language models to refine efficiency and reliability in multi-hop~QA.

\end{abstract}

\section{Introduction}

Large language models (LLMs) have recently demonstrated remarkable capabilities in open-domain question answering (QA), summarization, and various knowledge-intensive tasks~\cite{wang2023survey,kamalloo2023evaluating}. Relying solely on parametric knowledge embedded within an LLM poses challenges for correctness, interpretability, and scalability~\cite{kaddour2023challenges}. As queries become more complex or domain-specific, a single pass through the model may overlook vital details, leading to incomplete or inconsistent answers~\cite{zhu2021retrieving}. To tackle this, researchers have increasingly embraced the paradigm of Retrieval-Augmented Generation (RAG), where external corpora or knowledge bases are leveraged to supplement the LLM's internal representations~\cite{lewis2020retrieval}. 

While RAG has emerged as a promising approach to mitigating the limitations of purely parametric knowledge, simply retrieving or concatenating external evidence without a coherent strategy can still lead to fragmented reasoning and opaque answers~\cite{lewis2020retrieval,guu2020retrieval}. Recent work has focused on Knowledge Graph RAG pipelines, such as GraphRAG, where evidence is organized into a single graph for improved interpretability~\cite{edge2024local,soman2024biomedical,guo2024lightrag,kang2023knowledge}. Existing GraphRAG methods often rely on ad-hoc strategies that fail to exploit the full potential of multi-hop reasoning, especially when queries demand traversing multiple interconnected pieces of information.

Improving these limitations requires overcoming several hurdles. Specifically, naively performing one-shot retrieval for multi-hop questions can lead to either missing critical details or overwhelming the model with superfluous information~\cite{jiang2024retrieve}. Such a lack of selectivity compromises the clarity of intermediate reasoning steps by making it difficult to identify and integrate only the most relevant evidence~\cite{luo2025causal,zhu2025mitigating}. Moreover, even when a graph structure is introduced to enhance explainability, it often remains static or fails to synchronize with the evolving inference process~\cite{xu2021dynamic}. If nodes and edges are not updated at each intermediate step, the resulting graph can become cluttered or disconnected, creating gaps or redundancies that obscure the chain of reasoning~\cite{fang2019hierarchical}. These deficiencies are particularly evident in tasks that involve iterative or multi-turn queries, where previously retrieved knowledge must be revisited, refined, and logically connected to newly discovered evidence~\cite{sun2019pullnet}. Without a systematic way to incorporate these ongoing updates, key insights may remain isolated or overshadowed by irrelevant details, undermining the explainability and reliability of complex, multi-step reasoning.

The above pitfalls of fragmented evidence aggregation, unwieldy graph updates, and limited multi-turn support motivate our \textbf{StepChain GraphRAG} framework, which interleaves a \textbf{breadth-first search (BFS) Reasoning Flow (BFS-RF)} decomposition with explicit graph-based maintenance of intermediate results. Unlike conventional GraphRAG methods, which rely on static or shallow graph linking, StepChain GraphRAG dynamically updates the graph alongside each sub-question, allowing relevant evidence to be integrated in a principled manner~\cite{edge2024local}. Each sub-question triggers a targeted retrieval step, after which newly discovered evidence is inserted into the global graph via carefully defined node and edge creation rules. This mitigates redundancy by reusing previously discovered information and provides a transparent record of partial inferences that can be seamlessly revisited or refined. 

\begin{figure*}[t]
  \includegraphics[width=1\linewidth]{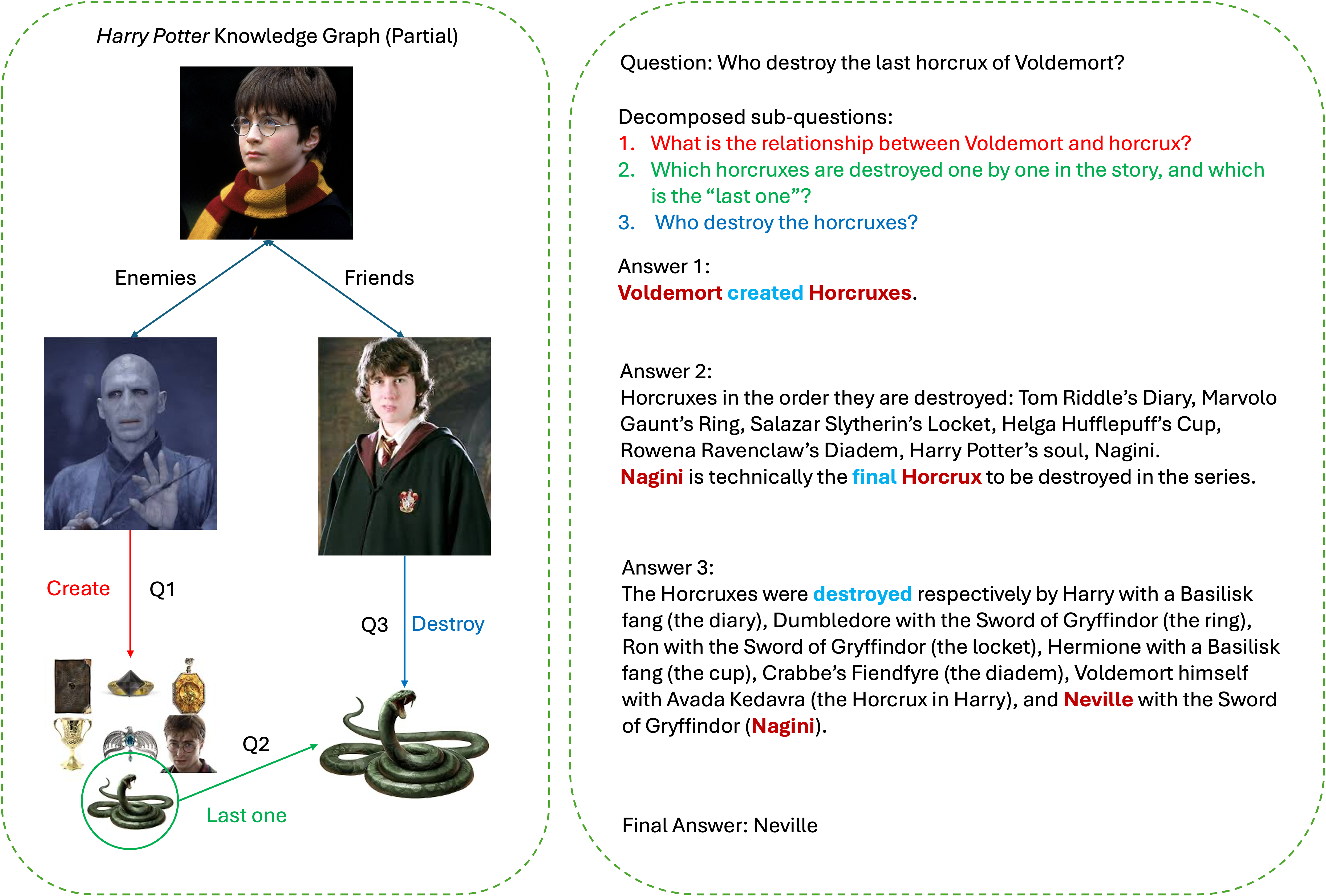}
  \caption{\textbf{Illustration of 
  StepChain GraphRAG
  applied to a Harry Potter example.} 
  On the left, a partial knowledge graph encodes entities (e.g., Voldemort, Nagini) and relationships (``create,'' ``destroy'') derived via entity extraction and linking. On the right, our system decomposes the user's query, ``Who destroys the last Horcrux of Voldemort?'', into sub-questions about (1) how Horcruxes relate to Voldemort, (2) which Horcrux is the final one, and (3) who destroys it. A BFS-based graph traversal gathers multi-hop evidence chains (e.g., ``Voldemort $\to$ creates $\to$ Horcruxes,'' ``Nagini $\to$ final Horcrux,'' ``Neville $\to$ destroys $\to$ Nagini''), and partial answers from each sub-question are merged to form the conclusion: \textbf{Neville} is the one who destroys Voldemort's last Horcrux.}
  \label{fig:example}
\end{figure*}

The following are the key contributions of this paper:
\begin{itemize}
    \item \textbf{Incremental Graph Updating:} We propose an incremental graph augmentation mechanism that synchronizes the knowledge graph with every round of newly retrieved evidence, ensuring the reasoning context remains consistently up‑to‑date throughout the multi‑hop process.
    \item \textbf{Iterative Multi-hop Reasoning:} We combine our BFS-RF with expansions over the current graph to tackle queries requiring multiple rounds of inference. Each sub-question triggers a targeted BFS traversal that uncovers cross-textual dependencies without overwhelming the model with extraneous content, thus delivering a focused and efficient retrieval process.
    \item \textbf{Enhanced Explainability:} By making intermediate results and their relationships explicit in a graph, our framework creates a more transparent and debuggable multi-hop QA pipeline than prior GraphRAG systems. The BFS-RF decomposition yields well-defined evidence paths at each step, allowing users and downstream modules to trace how partial inferences are built up from distinct pieces of retrieved information.
\end{itemize}
By systematically capturing and reusing these intermediate insights, StepChain GraphRAG alleviates information overload, delivers enhanced transparency in complex reasoning, and offers a debuggable pipeline for multi-hop QA. Through this explicit interplay between BFS-RF and dynamic graph expansion, our framework maintains an up-to-date, coherent knowledge structure that improves clarity and allows robustness in iterative or multi-turn query scenarios. 
Concretely, 
StepChain GraphRAG improves over the strongest baseline (HopRAG) by +1.70/+0.48 EM/F1 on MuSiQue, +1.30/+2.46 on 2WikiMultiHopQA, and +4.70/+3.44 on HotpotQA, yielding an average gain of +2.57 EM and +2.13 F1. 
We illustrate the end-to-end reasoning on an example in Fig.~\ref{fig:example}.

\section{Related Work}

\subsection{Stepwise Reasoning for Multi-hop QA}
Recent research on multi-hop QA has increasingly focused on explicit stepwise reasoning, ensuring that models break down complex queries into more manageable steps. Early methods introduced a pipeline that identifies simple queries and incrementally gathers evidence, demonstrating that step-by-step decomposition can improve accuracy and interpretability \cite{talmor2018web,min2019multi}. More recent approaches leverage LLMs by prompting them to articulate each reasoning step, often referred to as the ``chain-of-thought (COT)'' \cite{wei2022chain,kojima2022large}. 

These COT strategies can operate either in few-shot or zero-shot settings, and additional techniques, e.g., self-consistency, have been developed to mitigate the variability of different reasoning paths \cite{wang2022self}. A related idea is to guide the model through a structured, incremental solving process, by addressing the simplest parts of the problem first \cite{zhou2022least} or interleaving tool usage with model reasoning \cite{yao2023react}. Collectively, these studies highlight how prompting LLMs with intermediate steps can unlock powerful multi-hop reasoning capabilities, underscoring the importance of fine-grained decomposition to tackle complex queries.

\subsection{Retrieval-Augmented Generation}
A complementary research direction explores how to augment stepwise reasoning with external knowledge, giving rise to RAG models. Instead of relying solely on the language model’s parametric memory, these systems retrieve relevant passages from a large text corpus or knowledge base \cite{lewis2020retrieval,guu2020retrieval}. Once retrieved, the passages are fused into the generation process, often through specialized architectures that can handle multiple candidate documents \cite{izacard2020leveraging}. Scaling up these retrieval-centric systems has enabled near state-of-the-art performance in both fully supervised and few-shot open-domain QA \cite{izacard2023atlas}, emphasizing that external knowledge can bridge gaps in the model’s internal representation and reduce hallucinations. As multi-hop tasks grow in complexity, RAG methodologies become increasingly vital in ensuring that sufficient information is accessible to the reasoning module.

\subsection{Graph-based Retrieval and Reasoning}
Beyond plain text, there has been growing interest in leveraging \emph{graph structures} to organize and integrate knowledge for multi-hop reasoning. Some approaches build entity graphs from retrieved passages, propagating information across nodes to establish consistent chains of evidence \cite{qiu2019dynamically,ding2019cognitive,fang2019hierarchical}. Others utilize existing knowledge graphs or hyperlink graphs, treating multi-hop QA as a path-finding problem \cite{asai2019learning}. More recent endeavors incorporate domain-specific knowledge graphs directly into the retrieval pipeline, enhancing the model’s awareness of structured relationships among entities \cite{xu2024retrieval,zhu2024structugraphrag,tan2024paths}. Harnessing these relational cues, graph-based retrieval can improve both the interpretability and the accuracy of multi-hop reasoning.

In \cite{edge2024local}, a GraphRAG approach was proposed for query-focused summarization, which constructs a graph to integrate both local and global contexts from retrieved documents. By organizing evidence in a graph and capturing relational dependencies, their method demonstrates how a graph-centric strategy can effectively handle information fusion and improve the coherence of summarization. Although their work centers on summarization rather than QA, the principle of structuring retrieved evidence into interconnected graph nodes is readily transferable to multi-hop QA settings, where stepwise reasoning stands to benefit from explicit relational representations.

\subsection{Summary}

Despite recent progress in multi-hop QA, most approaches rely on static or ad-hoc expansions of knowledge graphs, limiting their capacity to handle iterative queries or incorporate newly uncovered information. We address these gaps with our StepChain GraphRAG framework, which combines question decomposition and BFS-RF with dynamic graph maintenance. This unified pipeline dynamically inserts new evidence at each sub-question, refining the knowledge graph in real time. The result is a more transparent, debuggable process for multi-hop QA by fully exploiting both text-based retrieval and graph-structured insights.

\section{StepChain GraphRAG}

\subsection{Overall Architecture}

\begin{figure*}[t]
  \centering
  \includegraphics[width=1\linewidth]{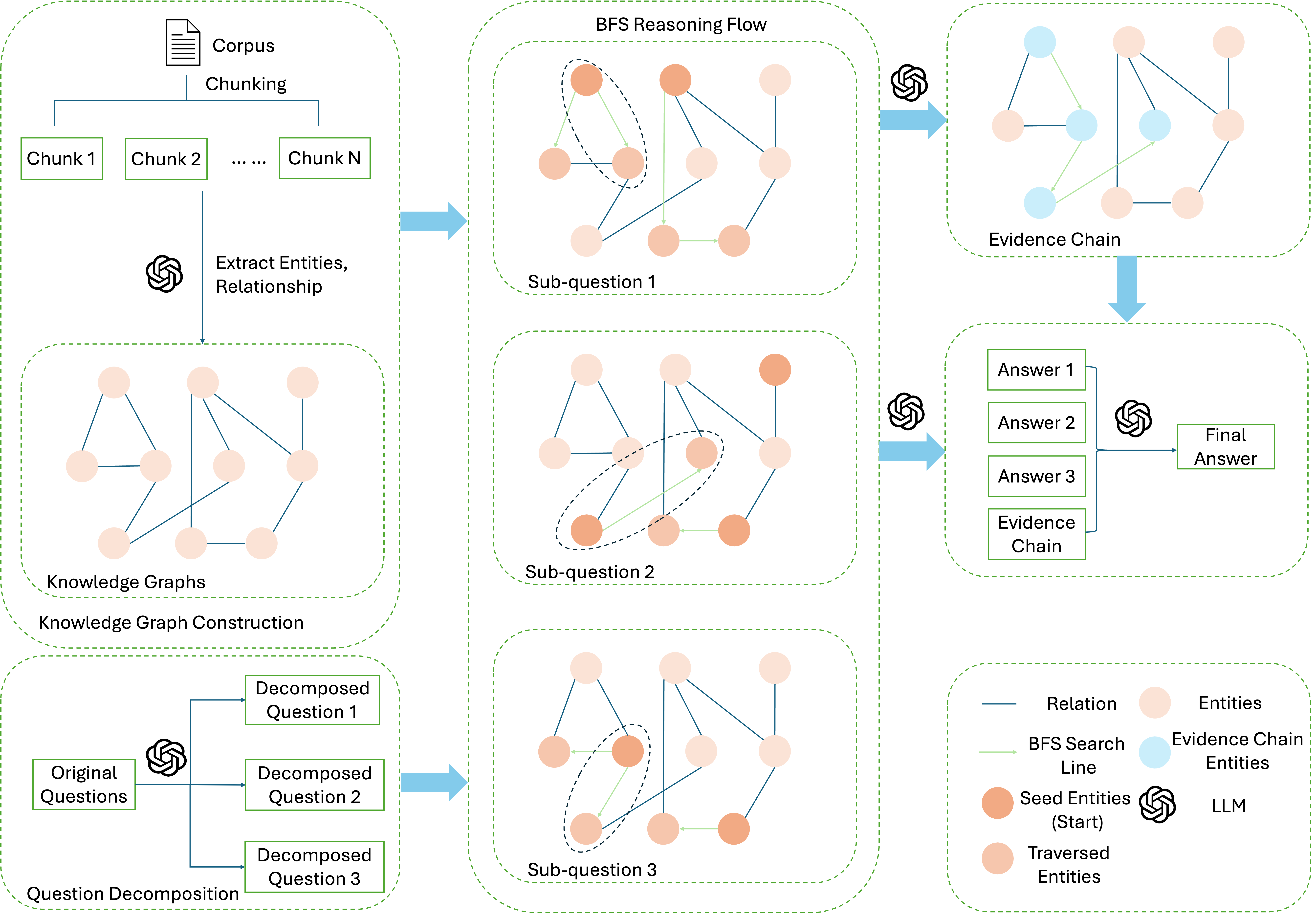}
  \caption{An overview of the StepChain GraphRAG pipeline. First, the corpus is split into chunks, and retrieved chunks are parsed on‑the‑fly to extract entities and relations and are upserted into a knowledge graph. Next, a complex question is decomposed into multiple sub-questions, each answered via BFS-RF that traverses the graph to find relevant entities and relations. The discovered evidence chains are combined to yield partial answers for each sub-question. Finally, these partial answers are synthesized by the LLM to produce the final, fully grounded response.}
  \label{fig:pipeline}
\end{figure*}

As shown in Fig. \ref{fig:pipeline}, we propose a multi-stage pipeline to handle complex multi-hop queries while preserving interpretability. First, raw texts are chunked into manageable segments and transformed incrementally into a knowledge graph that explicitly encodes entities and their relationships. This graph-based structure maintains a clear, traceable representation of how pieces of information interlink. Rather than processing an entire query in a single shot, we decompose it into smaller, more focused sub-questions. This decomposition prevents the model from conflating multiple logic threads at once, allowing it to target and retrieve only the most relevant nodes and edges.

Each sub-question is resolved through a BFS Reasoning Flow, where a breadth-first search systematically uncovers multi-hop evidence without overwhelming the model. Finally, the partial insights derived from each sub-question are merged into a single coherent answer, with a transparent trail illustrating how the evidence has been gathered and synthesized. By structuring the QA process into these steps, i.e., chunking and graph construction, question decomposition, BFS-based retrieval, and final synthesis, we reduce confusion, highlight key reasoning paths, and deliver more interpretable, high-fidelity answers than single-pass methods.

\subsection{Knowledge Graph Construction}
\label{sec:graph_construction}

We start by transforming the raw document corpus into a structured knowledge graph to support multi-hop QA. First, each document \(\tau_i\) in the corpus \(\{\tau_i\}_{i=1}^N\) is split into overlapping chunks for more manageable processing:
\begin{equation}
\label{eq:chunking}
D_i = \mathrm{Chunk}(\tau_i) = \bigl\{c_{i,1}, c_{i,2}, \ldots, c_{i,n_i}\bigr\},
\end{equation}
and all such chunks across the corpus form the following set: 
\begin{equation}
\mathcal{D} = \bigcup_{i=1}^{N} D_i.
\end{equation}

Next, we build the knowledge graph \(G=(V,E)\) by leveraging an LLM \(\mathcal{M}\) to identify named entities and their relationships in each retrieved chunk \(c_{i,j}\). In particular, the model extracts all entity mentions \(\bigl(e, \alpha_e\bigr)\), where \(\alpha_e\) may include attributes such as entity type or alias:
\begin{equation}
\mathrm{Extract}(c_{i,j}) \!
= \!\bigl\{(e,\, \alpha_e) \!\mid\! e \text{ appears in } c_{i,j}\bigr\}.
\end{equation}
Any two entities \((e_a,e_b)\) that co-occur in the same chunk lead to a relationship detection step,
\begin{equation}
r = \mathrm{Link}(e_a,\, e_b,\, c_{i,j}),
\end{equation}
producing an edge \(\bigl(e_a \xrightarrow{r} e_b\bigr)\) in \(G\). By explicitly capturing entities and their relations in graph form, we 
construct 
multi-hop paths that might span multiple documents or distant passages.

Note that, for the consideration of cost, we do not pre-parse the whole corpus into a graph. Only passages retrieved during reasoning are parsed into nodes/edges and upserted into \(G\) (see Sec.~\ref{sec:inc_graph}; for the first sub-question \(q_1\), we retrieve \(r\) passages from \(\mathcal{I}\) to initialize \(G_0\) with \(V_0\)). The entire corpus remains searchable in the global index \(\mathcal{I}\).
This chunking reduces noise by limiting each segment to a smaller context window, yet retains enough local coherence for accurate entity extraction and relationship detection. 
It also helps avoid exceeding LLM token constraints during subsequent analysis.

Finally, we optionally group highly interconnected entities via the Leiden community detection algorithm~\cite{traag2019louvain}:
\begin{equation}
\label{eq:community_detection}
\mathcal{C}_G = \mathrm{Cluster}(G) = \{C_1, C_2, \dots, C_K\}.
\end{equation}
Each community \(C_k \subseteq V\) can be summarized into a concise text snippet \(S_k\) with the LLM, which helps organize large graphs into topic-focused subgraphs. By capturing essential relationships in a structured form and optionally clustering them, we establish an explicit knowledge representation that better supports multi-hop reasoning.


\subsection{Question Decomposition}
\label{sec:question_decomposition}

Complex, multi-faceted user queries \(q\) often conflate multiple objectives when being handled in a single pass, leading to confusion and incomplete answers. To mitigate this risk, we introduce a specialized question decomposition stage, defined as
\begin{equation}
\label{eq:question_decompose}
\{q_1, q_2, \ldots, q_m\} = \mathrm{Decompose}(q),
\end{equation}
where \(\mathrm{Decompose}(q)\) splits the original query \(q\) into smaller or more targeted sub-questions. Each sub-question \(q_i\) isolates a particular logical component, hidden assumption, or sequential dependency. In practice, decomposition often employs carefully designed prompt templates that guide the LLM to identify distinct facets (e.g., an entity definition vs.\ its causal relationships).

This approach offers two advantages. On the one hand, it sharpens the focus of each retrieval step, ensuring that each sub-question draws upon only the most relevant evidence. On the other hand, it provides clearer interpretability by assigning each reasoning segment to a separate, traceable module. Unlike single-pass QA~\cite{kojima2022large,wei2022chain}, which risks blending multiple reasoning threads into one pass, our decomposition-based method produces reliable final answers and a transparent trail of how intermediate insights are derived.

\subsection{BFS-RF with Evidence Chain Provision}
\label{sec:bfs_reasoning}

Our framework designs BFS-RF to gather multi-hop evidence for each sub-question, reflecting the intuition that relevant information in a graph may lie several edges away from the starting point. By systematically expanding the search at each depth, BFS-RF mitigates the risk of missing crucial but indirectly connected nodes. Moreover, the evidence chains produced along each BFS path offer a clear, step-by-step audit trail of how specific entities and relations lead to partial answers, improving both interpretability and debuggability.

After decomposing the original query into sub-questions \(\{q_1,\ldots,q_m\}\), we retrieve the top-\(k\) entities in \(V\) whose embeddings (from a transformer or the LLM) best match each sub-question \(q_j\). Let \(\{s_1,\ldots,s_k\}\) denote these seed entities. We perform a breadth-first search out to depth \(h\), as given by
\begin{equation}
\label{eq:bfs}
\mathrm{BFS}(s_u, h) = \bigl\{ v \in V \;\big|\; \mathrm{dist}(s_u,v) \le h \bigr\},
\end{equation}
where \(\mathrm{dist}(s_u,v)\) is the shortest path distance in the knowledge graph \(G\). 
We also collect the exact path structures, as given by
\begin{equation}
\Pi_{s_u} \!\!=\!\! \{\pi \!\mid\! \pi \colon s_u \!\to \!\cdots\!\to\! v, \mathrm{dist}(s_u,v)\!\le\! h\},
\end{equation}
thus tracking how each reachable node links back to the seed entity. Each path \(\pi \in \Pi_{s_u}\) is translated into an ``evidence chain,'' for instance, \texttt{EntityA -- (relationX) --> EntityB -- (relationY) --> EntityC}, capturing not just which entities matter but how they connect. 
We concatenate these chain descriptions into a contextual string \(C_{q_j}\):
\begin{equation}
\label{eq:concatenate_context}
C_{q_j} = \bigl\|\mathrm{Desc}(\pi)\bigr\|_{\pi \in \bigcup_{u=1}^k \Pi_{s_u}},
\end{equation}
where \(\|\cdot\|\) denotes string concatenation. For \(j=1\), \(V\) denotes \(V_0\) obtained in the cold-start.
Feeding \(C_{q_j}\) and \(q_j\) into the LLM \(\mathcal{M}\) yields a partial answer \(A_j\) grounded in multi-hop evidence. 

The design behind BFS-RF and evidence chains rests on key objectives: capturing multi-hop links by examining nodes up to depth \(h\), where indirect or cross-textual dependencies might reside; providing a transparent record of nodes and edges observed, which aids error analysis or further refinement; and maintaining a targeted focus on each sub-question rather than scattering attention across many unrelated paths. These features, combined with the question decomposition, supply a structured and interpretable solution for multi-faceted, multi-hop queries.

Notably, in this paper, “retrieval on the graph” means that \(G\) constrains BFS expansions and provides path-level context; the actual passages are always fetched from the global index \(\mathcal{I}\).

\subsection{Incremental Graph Augmentation}
\label{sec:inc_graph}


After answering sub‑question \(q_j\), we query the global index \(\mathcal{I}\) conditioned on the current frontier \(F_j\) (entities/relations already visited) and obtain a small batch of passages \(\mathcal{D}^{\text{new}}\).
These passages are parsed on‑the‑fly to extract unseen entities and relations \((V^{\text{new}},E^{\text{new}})\), which are then upserted into the global graph \(G\):
\[
  G.\text{add\_nodes\_from}(V^{\text{new}}), \quad
  G.\text{add\_edges\_from}(E^{\text{new}}).
\]
This lazy regime reconciles the two descriptions: documents are retrieved from \(\mathcal{I}\), while \(G\), possibly incomplete, steers BFS expansions and records path‑level evidence chains.
This retrieve-and-refresh loop keeps the knowledge graph synchronized with the latest evidence, ensuring that every subsequent BFS traversal operates over an up-to-date context while allowing earlier nodes/edges to be enriched as information accrues.

Moreover, Recall is preserved by searching \(\mathcal{I}\) over the full corpus; precision is enforced via frontier‑conditioned queries, dense+BM25 re‑ranking, a minimum textual‑support threshold before adding edges, and caps on BFS depth and frontier size. If \(V\) becomes empty, we directly retrieve \(r\) passages from \(\mathcal{I}\) to re-seed before rerunning BFS.

\subsection{Merging Results and Final Model Output}
Once each sub-question $q_j$ has a partial answer $A_j$, we combine these intermediate findings into 
a more holistic view of $q$. Specifically, we take $\{A_1,\ldots,A_m\}$ and the optional community 
summaries $\{S_1,\ldots, S_K\}$, if relevant, and pass them to the LLM for a higher-level synthesis:
\begin{equation}
A_{\mathrm{merge}}\!\! =\!\! \mathcal{M}\Bigl(\!\{A_1,\!\ldots\!,A_m\},\,\{S_1,\!\ldots\!,S_K\}\!\Bigr).
\end{equation}

Each sub‑question retrieves its own top‑$k$ seed entities, whereas the community
summaries $\{S_1,\ldots,S_K\}$ are computed once on the global graph and
shared by all sub‑questions.
Here, $\mathcal{M}$ resolves potential overlaps or inconsistencies across sub-answers, integrates 
community-level context, and generates a coherent summary of how each sub-question informs the final solution. 

Finally, we feed the merged output $A_{\mathrm{merge}}$ back into the LLM alongside the original query 
$q$, i.e.,
\begin{equation}
\label{eq:final_answer}
A_{\mathrm{final}} = \mathcal{M}\bigl(q \,\|\, A_{\mathrm{merge}}\bigr),
\end{equation}
thus producing a comprehensive response that reflects both the local textual information (at the chunk 
level) and the global topological insight (at the graph level). This final answer can undergo optional 
post-processing (e.g., formatting to match user preferences, additional fact-checking, or 
filtering sensitive content) before delivery to the user.

Overall, by emphasizing question decomposition, BFS-based retrieval of evidence paths, and a 
two-tiered merging strategy, StepChain GraphRAG targets the shortcomings of 
purely single-turn QA systems.~\cite{wei2022chain,lewis2020retrieval}
It allows transparency in how answers are formed, leverages multi-hop connections, and scales 
well to queries that require crossing multiple documents or reasoning over distant entities.

\begin{table}[t!]
\centering
\label{tab:main_results}
\begin{tabular}{lcccccccc}
\toprule
\multirow{2}{*}{Method} 
 & \multicolumn{2}{c}{MuSiQue} 
 & \multicolumn{2}{c}{2Wiki} 
 & \multicolumn{2}{c}{HotpotQA} 
 & \multicolumn{2}{c}{Average} \\
\cmidrule(lr){2-3}\cmidrule(lr){4-5}\cmidrule(lr){6-7}\cmidrule(lr){8-9}
 & EM & F1 & EM & F1 & EM & F1 & EM & F1 \\
\midrule
BM25    & 13.80 & 21.50 & 40.30 & 44.83 & 41.20 & 53.23 & 31.77 & 39.85 \\
BGE     & 20.80 & 30.10 & 40.10 & 44.96 & 47.60 & 60.36 & 36.17 & 45.14 \\
GraphRAG & 12.10 & 20.22 & 22.50 & 27.49 & 31.70 & 42.74 & 22.10 & 30.15 \\
RAPTOR   & 36.40 & 49.09 & 53.80 & 61.45 & 58.00 & 73.08 & 49.40 & 61.21 \\
SiReRAG  & 40.50 & 53.08 & 59.60 & 67.94 & 61.70 & 76.48 & 53.93 & 65.83 \\
HopRAG   & 42.20 & 54.90 & 61.10 & 68.26 & 62.00 & 76.06 & 55.10 & 66.40 \\
\cmidrule(lr){1-9}
Ours     & \textbf{43.90} & \textbf{55.38} & \textbf{62.40} & \textbf{70.72} & \textbf{66.70} & \textbf{79.50} & \textbf{57.67} & \textbf{68.53} \\
\bottomrule
\end{tabular}
\caption{QA performance (EM and F1) on MuSiQue, 2Wiki, and HotpotQA
with GPT-4o using top-20 passages, plus their averages.}
\end{table}

\begin{table}[t]
    \centering
    \label{tab:llm_ablation}
    \begin{tabular}{lcc}
    \toprule
    \textbf{LLM} & \textbf{EM} & \textbf{F1} \\
    \midrule
    Llama\,3.3 (70B) & 54.00 & 68.77 \\
    Qwen\,2.5 (72B)  & 59.30 & 71.33 \\
    GPT-4o     & \textbf{66.70} & \textbf{79.50} \\
    \bottomrule
    \end{tabular}
    \caption{Ablation on HotpotQA with various LLMs.}
\end{table}

\begin{table*}[ht]
    \centering
    \caption{Ablation study with GPT-4o on MuSiQue, 2Wiki, and HotpotQA.
    We report EM and F1 for each dataset, as well as the averaged EM/F1 across all three.}
    \label{tab:ablation_study}

    \setlength{\tabcolsep}{4pt}
    \renewcommand{\arraystretch}{1.12}

    \resizebox{\textwidth}{!}{%
    \begin{tabular}{lcccccccc}
        \toprule
        \multirow{2}{*}{\textbf{Method}}
            & \multicolumn{2}{c}{\textbf{MuSiQue}}
            & \multicolumn{2}{c}{\textbf{2Wiki}}
            & \multicolumn{2}{c}{\textbf{Hotpot}}
            & \multicolumn{2}{c}{\textbf{Average}} \\ 
        \cmidrule(lr){2-3}\cmidrule(lr){4-5}\cmidrule(lr){6-7}\cmidrule(lr){8-9}
            & \textbf{EM} & \textbf{F1}
            & \textbf{EM} & \textbf{F1}
            & \textbf{EM} & \textbf{F1}
            & \textbf{EM} & \textbf{F1} \\
        \midrule
        GPT-4o                                          & 10.80 & 18.52 & 21.20 & 25.68 & 30.30 & 40.41 & 20.77 & 28.20 \\
        GraphRAG Baseline                               & 12.10 & 20.22 & 22.50 & 27.49 & 31.70 & 42.74 & 22.10 & 30.15 \\
        Question Decomposition Baseline                 & 21.50 & 31.40 & 43.90 & 47.06 & 43.60 & 58.94 & 36.33 & 45.80 \\
        GraphRAG + Question Decomposition               & 36.40 & 49.09 & 50.80 & 61.45 & 56.90 & 70.08 & 48.03 & 60.21 \\
        GraphRAG + Reasoning                            & 39.60 & 50.71 & 52.20 & 63.80 & 58.60 & 72.21 & 50.13 & 62.24 \\
        GraphRAG + Reasoning + Question Decomposition   & 43.90 & 55.38 & 62.40 & 70.72 & 66.70 & 79.50 & 57.67 & 68.53 \\
        \bottomrule
    \end{tabular}%
    }
\end{table*}

\section{Experiment}

\subsection{Experimental Setup}

To demonstrate the effectiveness of StepChain GraphRAG, we select three representative multi-hop QA datasets: MuSiQue~\cite{trivedi2022musique}, 2WikiMultiHopQA~\cite{ho2020constructing}, and HotpotQA~\cite{yang2018hotpotqa}. Using the same corpus as HippoRAG~\cite{gutierrez2024hipporag}, we choose 1,000 questions from each validation set of these three datasets. All questions require multi-hop reasoning across multiple paragraphs, making them suitable benchmarks for evaluating complex retrieval and reasoning techniques. 

We use the standard Exact Match (EM) and F1 scores to measure the accuracy of predicted answers compared to ground-truth labels. EM requires an exact string match with the gold answer; F1 rewards partial overlaps by considering token-level precision and Recall. These two metrics offer a balanced perspective on both strict correctness and partial completeness of the system's responses.

All experiments are conducted in Python with a graph-based retriever and BFS expansions. GPT-4o serves as the default QA model (for comparability with SOTA) unless otherwise noted. Chunks contain 1,200 tokens with 100-token overlap; BFS depth is 2. LLM calls are issued asynchronously with caching to reduce latency. All tests run on two RTX 6000 Ada GPUs using the same prompts and hyperparameters as 
prior retrieval-based work.~\cite{liu2025hoprag}

\subsection{Main Results and Comparison with SOTA}
Table 1 compares StepChain GraphRAG against an array of baselines on MuSiQue, 2Wiki, and HotpotQA, each evaluated with GPT-4o under a top-20 passage budget. Sparse (BM25)~\cite{robertson2009probabilistic} and dense (BGE)~\cite{karpukhin2020dense} retrievers provide initial reference points, while GraphRAG~\cite{edge2024local}, RAPTOR~\cite{sarthi2024raptor}, SiReRAG~\cite{zhang2024sirerag}, and HopRAG~\cite{liu2025hoprag} represent state-of-the-art structured retrieval. StepChain GraphRAG attains superior EM and F1 on all three datasets, indicating that our BFS-driven graph expansions effectively capture multi-hop dependencies. Across all tasks, an average EM/F1 of 57.67/68.53 is achieved, highlighting consistent gains even in the presence of distractors and diverse query types.

We investigate the robustness of our pipeline under different LLMs by focusing on HotpotQA. Table 2 compares the final performance obtained when we substitute GPT-4o with Llama\,3.3 (70B)and Qwen\,2.5 (72B). While Llama\,3.3 and Qwen\,2.5 both surpass 
earlier baselines \cite{grattafiori2024llama,yang2025qwen3}
, Qwen\,2.5 offers a slight edge, indicating nuanced differences in model architecture and training data. Notably, GPT-4o attains the highest EM and F1 (66.70 and 79.50), suggesting that model choice significantly affects multi-hop QA performance. Our results reinforce the synergy between advanced graph retrieval modules and robust generative models to handle multi-step reasoning effectively.

\subsection{Ablation Study}
We assess each component’s contribution to accuracy. Table~\ref{tab:ablation_study} shows monotonic gains across MuSiQue, 2Wiki, and HotpotQA as retrieval and reasoning are added. GPT‑4o (no retrieval) averages \textbf{20.77/28.20} (EM/F1); graph retrieval alone and question decomposition alone reach \textbf{22.10/30.15} and \textbf{36.33/45.80}. Integrating decomposition with the graph achieves \textbf{48.03/60.21}, adding explicit graph reasoning yields \textbf{50.13/62.24}, and combining decomposition+reasoning peaks at \textbf{57.67/68.53}, confirming the synergy of decomposition, graph retrieval, and chain‑of‑thought reasoning for multi‑hop QA.

\subsection{Inference time and latency analysis}
The end‑to‑end latency of the proposed approach is dominated by the LLM inference. GPT‑4o API takes approximately $80$s per query, due to network delays and service rate limits. Self‑hosted Qwen‑2.5‑72B and Llama‑3.3‑70B require approximately $90$s and $94$s, respectively (on two RTX 6000Ada). In contrast, retrieval, BFS traversal, and incremental graph updating together introduce less than $3$s overhead, showing that graph logic is not a bottleneck for throughput.  
The vast majority of the wall‑clock time is therefore spent on LLM inference. 
Since graph updates are executed in‑memory and streamed directly into the prompt, their cost grows modestly with the number of hops. As a result, practical latency improvements should focus on the LLM itself, including prompt caching, lightweight quantization, speculative decoding, or swapping in a smaller distilled checkpoint, rather than on the graph layer.

\section{Conclusion}
We have presented \textit{StepChain GraphRAG}, a framework that integrates stepwise COT reasoning with graph-based retrieval to tackle multi-hop QA. By iteratively updating a knowledge graph with each sub-question and partial answer, our approach clarifies the evidence chain, curbs information overload, and bolsters interpretability. Empirical results on HotpotQA indicate that such a structured pipeline can surpass ad-hoc retrieval methods, 
yet StepChain GraphRAG’s reliance on explicit graph construction imposes extra computational overhead and memory demands.

\section{Limitation and Future Work}
LLMs can hallucinate, allowing spurious facts to propagate through the pipeline and weaken answers. Additionally, in multi-hop datasets where sub-question \(q_j\) depends on \(q_{j-1}\), errors or uncertainty in early steps can make retrieval for \(q_j\) brittle. Our StepChain GraphRAG pipeline partially mitigates this issue via per-sub-question retrieval from the global index \(\mathcal{I}\) and explicit BFS evidence chains for interpretability. However, we do not quantify the residual impact here and plan to introduce uncertainty-aware re-decomposition and backtracking, as well as sub-question-level robustness metrics, in future work. Graph-centric retrieval struggles in specialized or rapidly changing domains, and GraphRAG imposes computational overhead at inference time. Future work will also reduce inference-time cost, deliver scalable updates to the graph, strengthen verification against hallucinations, and better handle incomplete or domain-specific data.
\bibliographystyle{splncs04}
\bibliography{mybibliography}





\end{document}